\documentclass{article}

\usepackage[eandd, final]{neurips_2026}

\usepackage[utf8]{inputenc} 
\usepackage[T1]{fontenc}    
\usepackage{hyperref}       
\usepackage{url}            
\usepackage{booktabs}       
\usepackage{amsfonts}       
\usepackage{nicefrac}       
\usepackage{microtype}      
\usepackage{xcolor}         

\usepackage{microtype}
\usepackage{graphicx}
\usepackage{subcaption}
\usepackage{hyperref}
\usepackage{amsmath}

\title{ShallowBench: Benchmarking Generative Drug Design Models on Shallow-Pocket Targets}

\author{
    Saket Reddy \quad Shiwei Liu \\
    University of Illinois - Urbana-Champaign \\
    \texttt{\{saketr3, shiweil2\}@illinois.edu}
}

\begin{document}

\maketitle

\begin{abstract}
While generative AI models have demonstrated remarkable success in structure-based drug design, they predominantly rely on deep binding pockets and struggle to sample effective ligands for challenging low-pocketability targets, such as the historically “undruggable” oncology targets KRAS and MYC. To address this gap, we introduce ShallowBench, a strictly curated benchmark of 5,780 shallow-pocket targets extracted from CrossDocked2020. By computing the difference between an Alpha Shape “lid” volume and the underlying protein atom voxel volume, we successfully isolated targets with low concavity while ensuring sufficient surface area for binding. Evaluating various state-of-the-art generative models reveals weaker predicted binding affinity on these low-concavity interfaces. ShallowBench therefore provides a rigorous benchmark for generative biology models and highlights the necessity of new architectural innovations or loss functions capable of navigating these challenging targets.
\end{abstract}

\section{Introduction}
Generative models such as $SE(3)$-equivariant diffusion and flow-matching neural networks have demonstrated remarkable proficiency in structure-based drug design (SBDD) \citep{schneuing2023diffsbdd}, successfully producing chemically valid, high-affinity molecules. However, the success of these models is predicated on the presence of deep, structurally-defined binding pockets. Deep cavities provide clear generation constraints and expansive surface areas for robust Van der Waals interactions, effectively anchoring the generated coordinates \citep{corso2022diffdock}. 

Consequently, an important vulnerability in the current SBDD paradigm remains largely unaddressed: the generation of ligands for shallow or intrinsically disordered protein surfaces. Many of the most highly sought-after therapeutic targets in oncology, such as KRAS and Myc, lack traditional, high-concavity binding pockets.\citep{kessler2019drugging, llombart2022therapeutic}. Ligands attempting to bind to these flat interfaces face higher bulk-solvent competition and lack well-defined structural enclosures. Compounding this issue is a persistent training and evaluation bias. Standard benchmark datasets such as CrossDocked2020 \citep{francoeur2020crossdocked} and PDBbind \citep{liu2015pdbbind} are dominated by deep-pocket targets. As a result, state-of-the-art SBDD models learn skewed distributions \citep{gavali2024dataset}. 

Because the field lacks a large, dedicated benchmark for shallow targets, the performance degradation of these models on flat surfaces remains unclear, hindering the development of architectures capable of tackling non-traditional binding sites. To bridge this gap, we introduce ShallowBench, a specialized benchmark rigorously curated to evaluate generative drug design models on challenging, low-concavity protein targets. 

Our contributions are as follows:
\begin{itemize}
    \item \textbf{Data Curation and Benchmark Set:} We develop a two-step volumetric approach using an Alpha Shape ``lid'' calculation to effectively isolate 5,780 shallow-pocket targets from the CrossDocked2020 dataset, ensuring low concavity while maintaining sufficient surface area for binding. 
    \item \textbf{Fine-Tuning Dataset:} We provide strictly split training (4,995 targets) and testing (785 targets) sets using 30\% sequence identity clustering to prevent leakage, providing a valuable resource for fine-tuning generative models. \textbf{We release the full, train, test, and control datasets on \href{https://huggingface.co/datasets/SaketR1/shallowbench/tree/main}{Hugging Face}.} (\url{https://huggingface.co/datasets/SaketR1/shallowbench/tree/main}). 
    \item \textbf{Model Evaluation and Benchmarking Findings:} We evaluate state-of-the-art generative SBDD models on ShallowBench, revealing a systematic decline in predicted binding affinity across all evaluated architectures. We further expose vulnerabilities in other metrics when generating ligands for flat surfaces, such as a degradation in chemical validity for TargetDiff. These performance gaps demonstrate the need for new architectural innovations capable of navigating low-concavity targets.
\end{itemize}

\section{Background and Related Work}
\subsection{Structure-Based Drug Design}
Structure-based drug design (SBDD) has seen tremendous progress with the advent of generative AI, particularly with the introduction of equivariant diffusion models. These generative models have demonstrated remarkable success in consistently generating chemically valid and high-affinity ligands \citep{schneuing2023diffsbdd}. However, the current state-of-the-art relies heavily on deep pockets to generate effective ligand coordinates. Deep cavities intrinsically provide the necessary surface area for robust Van der Waals interactions and serve as well-defined anchor points for 3D coordinate generation. Furthermore, because these models are predominantly evaluated and trained on standard datasets dominated by high-concavity sites, they learn biased distributions \citep{gavali2024dataset}. 

\subsection{Low-Pocketability and Shallow-Pocket Targets}
While deep pockets are ideal for computational modeling, many highly sought-after therapeutic targets possess shallow or intrinsically disordered binding interfaces. For instance, KRAS, one of the most frequently mutated oncogenes, was historically deemed ``undruggable'' due to a lack of traditional binding cavities \citep{kessler2019drugging}. Similarly, Myc, which is altered in many human cancers, lacks a well-defined hydrophobic pocket and instead relies on an expansive flat surface for protein-protein interactions \citep{llombart2022therapeutic}. 

Generating ligands for these flat surfaces presents three primary challenges:
\begin{enumerate}
    \item \textbf{Solvent Exposure:} Ligands situated on flat surfaces face increased bulk-solvent competition. For example, hydrogen bonds and electrostatic interactions can be more simply disrupted by surrounding water molecules \citep{korolev2002competition}. 
    \item \textbf{Sparse Contact Area:} Flat interfaces provide minimal 3D constraints. Without the geometric bounds of a deep pocket, generative models can produce coordinates that ``float'' away from the protein \citep{dong2026flat}.
    \item \textbf{Training Data Bias:} Standard SBDD datasets used to train models are skewed towards high-concavity sites \citep{gavali2024dataset}. 
\end{enumerate}

\subsection{Existing Datasets}
The evaluation of SBDD models currently relies on standard datasets like CrossDocked2020 and PDBbind, which are dominated by deep pockets and therefore may mask the vulnerabilities of generative models on shallow surfaces \citep{gavali2024dataset}. The closest existing data curation efforts for shallow-pocket targets involve cryptic pocket datasets. For example, PocketMiner introduced a highly curated and verified set of 39 cryptic pockets \citep{meller2023predicting}. Similarly, CryptoBench provides a more expansive set of 1,107 cryptic structures \citep{skrhak2025cryptobench}. However, these datasets primarily focus on pockets that open up during molecular dynamics (MD) simulations rather than evaluating the generation of ligands for static, naturally shallow or flat surfaces. Therefore, a dedicated benchmark for static shallow pockets remains a critical gap in the field.

\section{Dataset Curation}
\begin{figure*}[t]
    \centering
    \includegraphics[width=14cm]{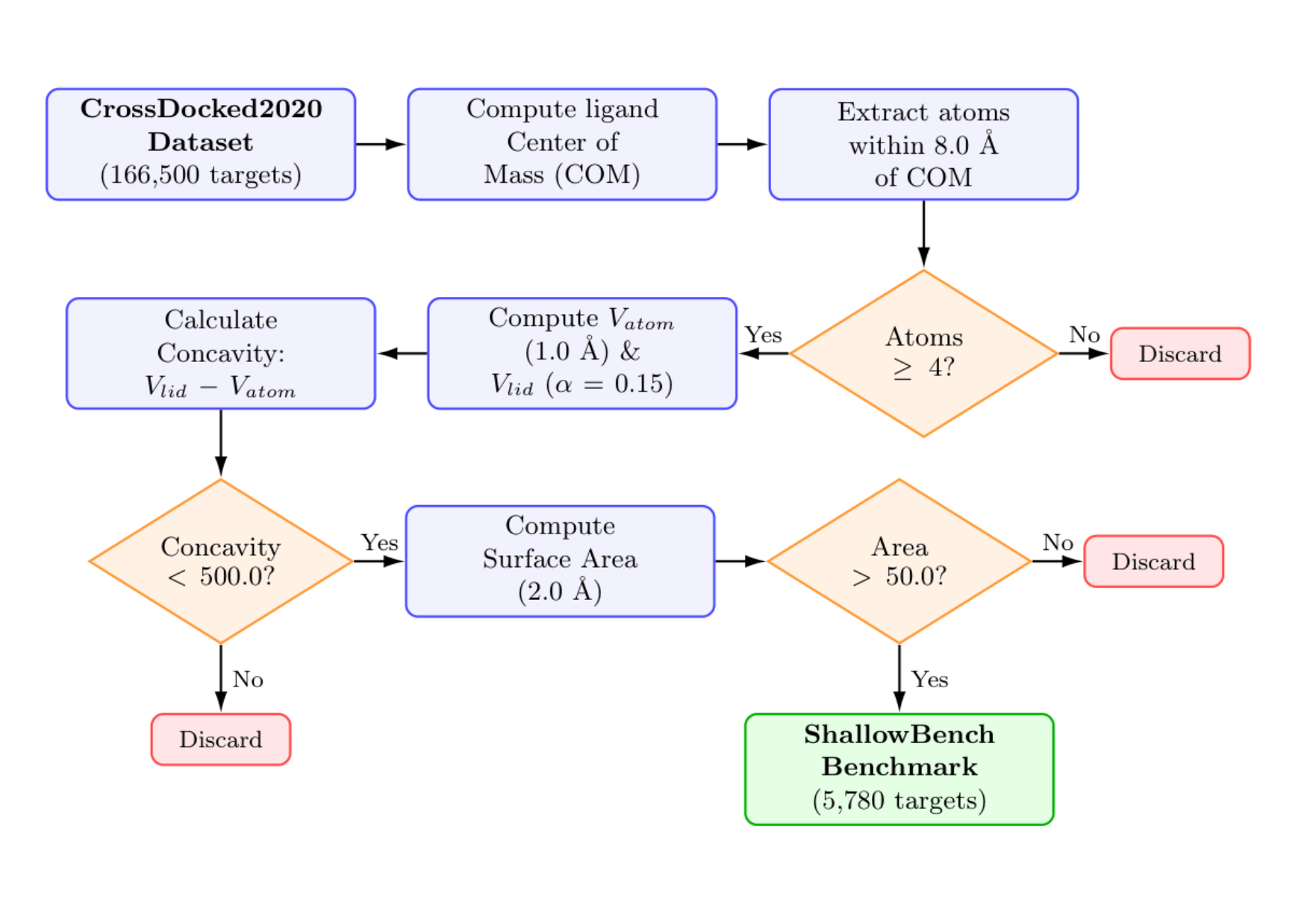}
    \caption{Data curation pipeline for isolating shallow-pocket targets in ShallowBench.}
    \label{fig:shallowbench_curation}
\end{figure*}

\subsection{ShallowBench Curation}
To evaluate generative models on realistic shallow targets, we curated a specialized benchmark dataset, ShallowBench, derived from the comprehensive CrossDocked2020 dataset \citep{francoeur2020crossdocked}. The curation process was designed to rigorously isolate protein interfaces that lack deep cavities while still possessing sufficient surface area for potential ligand binding. 

For each protein-ligand complex, we first defined the local binding environment by computing the center of mass (COM) of the native ligand. We extracted all protein atoms within an $8.0 \text{\AA}$ radius of this COM. Any interfaces containing fewer than 4 atoms were discarded to ensure validity. 

To quantify the concavity of a target's pocket, we developed a two-step volumetric approach. First, we established a baseline protein atom volume ($V_{atom}$) by mapping the extracted interface coordinates to a 3D voxel grid with a voxel size of $1.0 \text{\AA}$, calculating the total spatial volume strictly occupied by the protein atoms. Second, to measure the empty, targetable space directly above the binding surface, we generated an Alpha Shape mesh \citep{alphashape} over the interface coordinates utilizing an alpha parameter of $\alpha = 0.15$ (alpha parameter empirically chosen for producing best volume calculations after various testing). This mesh acts as a simulated ``lid'' enclosing the pocket, providing a bounding volume ($V_{lid}$). The concavity of the target surface was then defined as the difference between these two volumes:

$$ \text{Concavity} = V_{lid} - V_{atom} $$ 

\begin{figure*}[h]
    \centering
    \includegraphics[width=12cm]{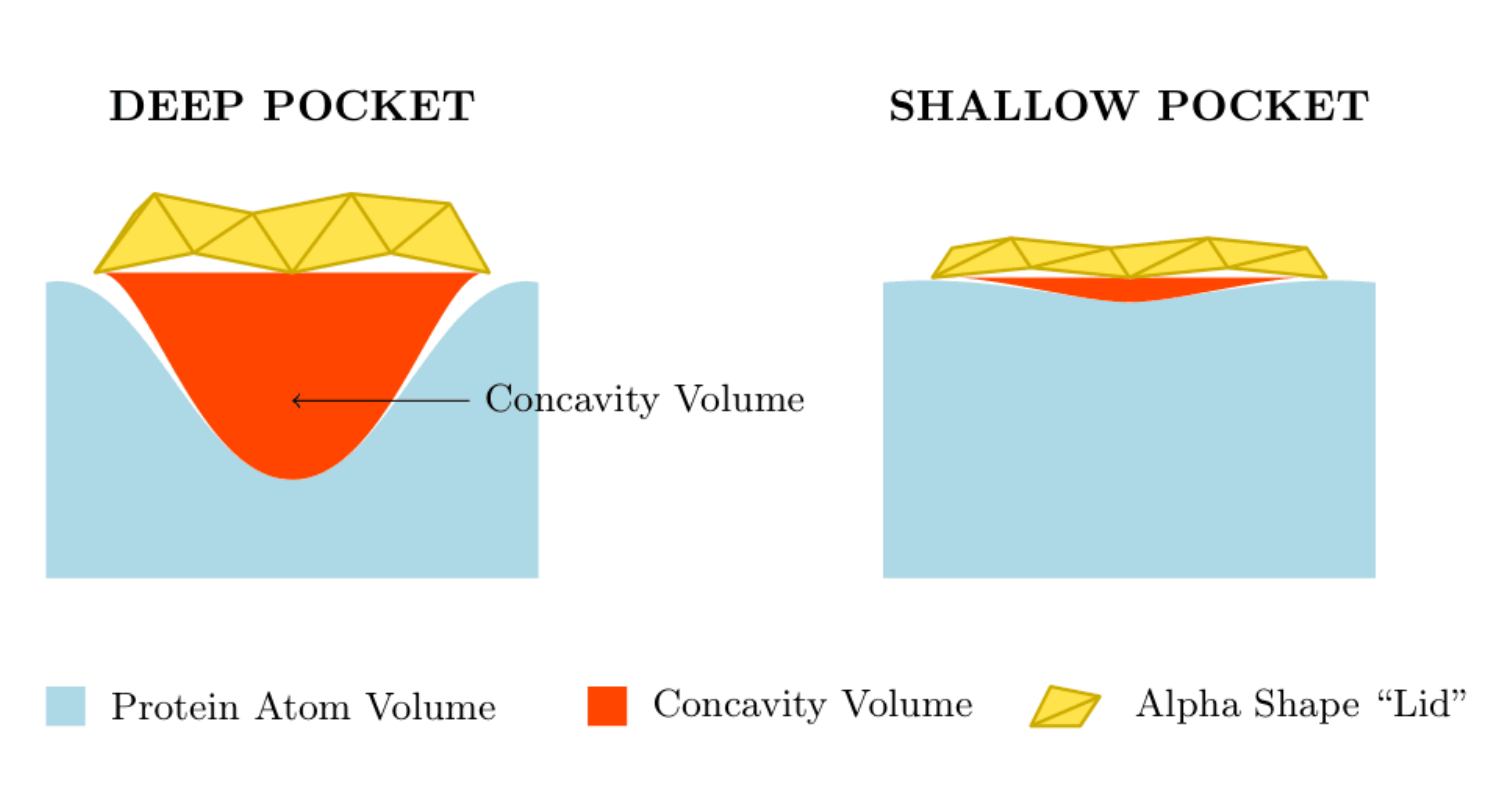}
    \caption{Comparison of deep pocket volume to shallow pocket volume.}
    \label{fig:pocket_volume_comparison}
\end{figure*}

A visualization of this calculation is shown in Figure~\ref{fig:pocket_volume_comparison}. To select shallow surfaces, we enforced a strict upper bound of $\text{Concavity} < 500.0$ Å³. To ensure that the selected shallow interfaces still offered enough physical space for binding, we calculated the surface area using a larger $2.0 \text{\AA}$ voxel grid and applied a lower bound threshold of $\text{Surface Area} > 50.0$ Å². This rigorous pipeline yielded 5,780 shallow-pocket targets, from a starting point of 166,500 total targets. A visualization of the entire curation pipeline is shown in Figure~\ref{fig:shallowbench_curation}. 

\subsection{ShallowBench Analysis}

\begin{table}[htbp]
    \centering
    \caption{Concavity and Surface Area Statistics}
    \vspace{0.2cm} 
    \begin{tabular}{l r r r r}
        \toprule
        \textbf{Metric} & \textbf{Mean} & \textbf{Median} & \textbf{Min} & \textbf{Max} \\
        \midrule
        Concavity (Å³)   & 361.07 & 385.15 & -0.54 & 500.00 \\
        Surface Area (Å²) & 272.68 & 280.00 & 56.00 & 584.00 \\
        \bottomrule
    \end{tabular}
    \label{tab:concavity_sa_stats}
\end{table}

\begin{figure*}[h]
    \centering
    \includegraphics[width=14cm]{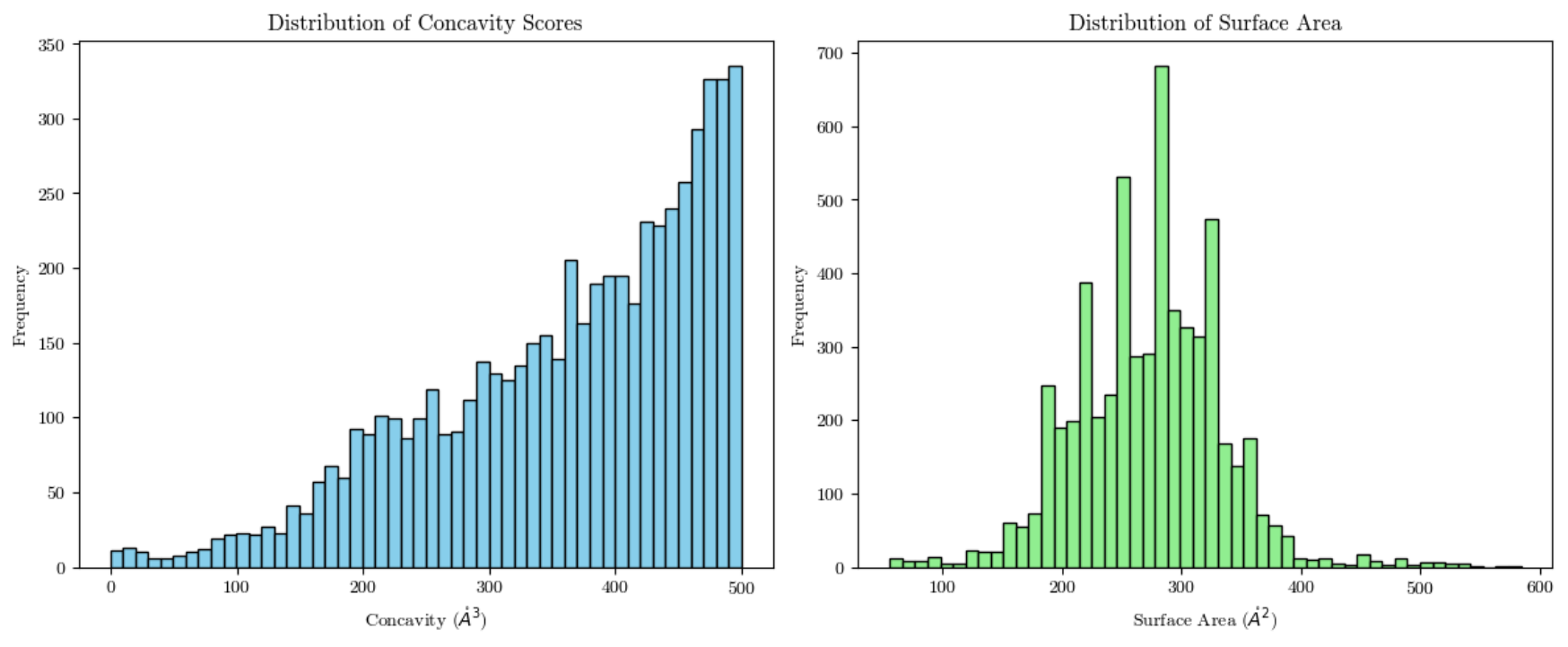}
    \caption{Distribution of concavity and surface area of targets in ShallowBench.}
    \label{fig:distribution_plots}
\end{figure*}

Table~\ref{tab:concavity_sa_stats} details summary statistics for the concavity and surface area of the targets in ShallowBench, and Figure~\ref{fig:distribution_plots} shows visualizations of the distributions. 

For concavity, the dataset exhibits a mean of 361.07 and a median of 385.15, ranging from a minimum of -0.54 up to the maximum threshold of 500.00. The concavity distribution is left-skewed, reflecting the data curation process which visually sliced off the left tail of the larger CrossDocked dataset to successfully isolate the low-concavity targets. 

For surface area, the dataset demonstrates a mean of 272.68 and a median of 280.00, bounded by a minimum of 56.00 and a maximum of 584.00. The surface area distribution forms a bell curve. 

The concavity upper bound of 500.00 was empirically chosen in the context of this surface area distribution. A threshold of 500 ensured low pocketability while still yielding the healthy, normal-like surface area distribution shown. This healthy distribution indicates that the concavity constraint successfully captures a distinct, physically coherent class of protein interfaces that, despite lacking deep pockets, still retain sufficient spatial area for potential ligand binding. 

\subsection{Train/Test Sequence Splitting}
To provide a resource for researchers who would like to fine-tune their models on shallow-pocket targets, we also split the 5,780 curated targets into train and test splits. 

To prevent data leakage between homologous proteins, we split targets based on sequence homology. We utilized the Protein Data Bank (RCSB) \citep{liu2015pdbbind} 30\% sequence identity clusters to group all corresponding PDB IDs. By uniformly distributing entire sequence clusters into their respective sets, we guaranteed that no target in the test set shares greater than 30\% sequence similarity with any protein in the training set, ensuring the test set is rigorous for researchers who engage in fine-tuning. This process resulted in a split of 4,995 training targets and 785 test targets. 

\subsection{Control Dataset}

\begin{figure*}[h]
    \centering
    \includegraphics[width=14cm]{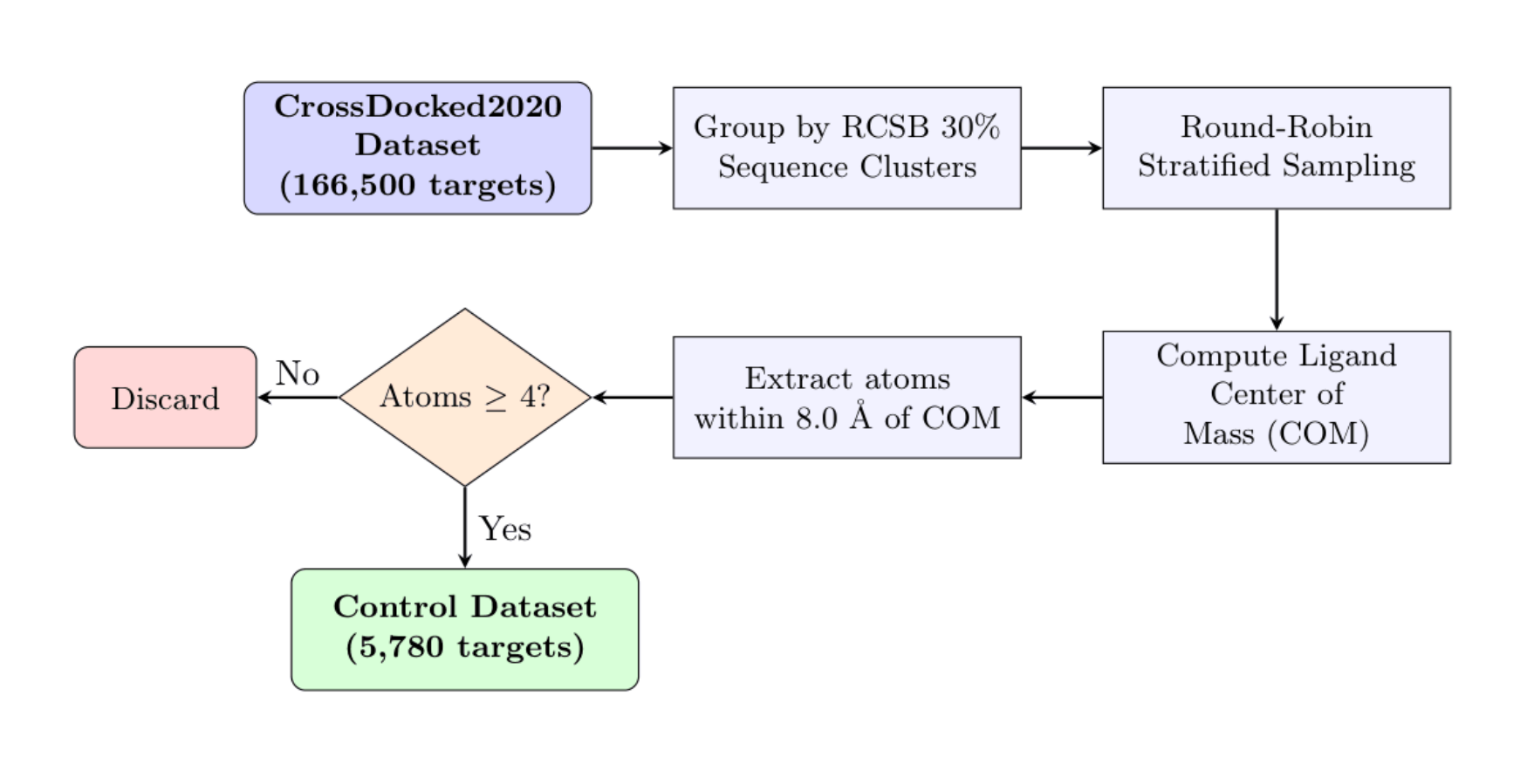}
    \caption{Data curation pipeline for the control dataset.}
    \label{fig:control_curation}
\end{figure*}

To establish a reliable baseline for evaluating the performance disparity of generative models on shallow-pocket targets, we curated a structurally diverse control dataset. We extracted exactly 5,780 targets from the standard CrossDocked2020 dataset, ensuring the control set mirrors the size of the ShallowBench benchmark. 

To maximize structural diversity and prevent the over-representation of a small number of protein families, we implemented a round-robin stratified sampling technique. Using the RCSB 30\% sequence identity clusters previously established in Section 3.3, we first mapped every available CrossDocked2020 complex to its corresponding sequence cluster. We then initialized a seeded randomization, shuffling both the overarching clusters and the constituent targets within them. By iterating through the sequence clusters and sampling one target from each cluster in a continuous round-robin sequence, we successfully collected 5,780 unique target indices. This stratification guarantees a very heterogeneous control subset that accurately reflects the broad distribution of CrossDocked2020.

For preprocessing, the selected targets were parsed from the CrossDocked LMDB environment. Mirroring the geometric constraints used for ShallowBench, we established the local binding environment by computing the native ligand's COM and extracting all underlying protein atoms within an $8.0 \text{ \AA}$ radius. To maintain structural validity and downstream compatibility, we discarded any interfaces containing fewer than 4 atoms. The final control dataset provides a rigorously sampled, structurally diverse baseline that is formatted identically to ShallowBench, allowing for a robust, one-to-one comparative evaluation of regular vs. shallow-pocket performance. An illustration of the pipeline is shown in Figure~\ref{fig:control_curation}. 

\section{Evaluation}
\subsection{Metrics}
To assess the capability of current models to navigate shallow topologies, we employed a suite of robust evaluation metrics. We evaluated these metrics across the entire 5,780 benchmark dataset; no fine-tuning was performed.  
\begin{itemize}
    \item \textbf{Chemical Validity:} We measured the proportion of generated molecules that pass fundamental valence, aromaticity, and sanitization checks using RDKit's \texttt{Chem.SanitizeMol} \citep{rdkit}.
    \item \textbf{Mean QED:} We used the Quantitative Estimate of Druglikeness, scaled from 0 to 1, to measure whether the generator maintains realistic pharmaceutical properties \citep{bickerton2012quantifying}. 
    \item \textbf{Vina Affinity:} We utilized the predicted binding energy ($\text{kcal/mol}$) calculated via AutoDock Vina, operating within a $20.0 \text{\AA}$ bounding box centered around the native ligand's COM \citep{trott2010autodock}.
    \item \textbf{Shape Complementarity (Sc):} We calculated Sc via the SCASA algorithm \citep{lawrence1993shape}. This metric measures the geometric fit between the ligand and protein. Because our data curation strictly filtered for shallow surfaces, Sc is inherently expected to be substantially lower than on deep pockets. We utilized Sc not as an optimization target, but to serve as a negative control confirming our filtering was successful, and to reveal whether generative models can establish any geometric traction on flat planes. 
\end{itemize}

\subsection{Models}
To evaluate performance on shallow surfaces, we benchmarked various frontier conditional structure-based drug design (SBDD) models.
\begin{itemize}
    \item \textbf{DiffSBDD}: DiffSBDD \citep{schneuing2023diffsbdd} is an SE(3)-equivariant conditional diffusion model, designed to generate novel ligands conditioned on pre-determined protein pockets by operating directly on continuous 3D coordinates. During its training and generation phases, the model progressively denoises both categorical atom types and 3D spatial coordinates. By utilizing an equivariant graph neural network (GNN), the architecture preserves translation and rotational symmetries. 

    \item \textbf{SimpleSBDD}: SimpleSBDD \citep{karczewski2025model} is a conditional model that was developed to investigate and counter the over-parameterization and limitations typically found in standard GNN architectures. Instead of relying on complex generative networks, it employs a streamlined metric-aware approach that learns a surrogate for binding affinity to infer an unlabeled molecular graph. A unique aspect of its training is that it optimizes for labels conditioned on this graph alongside molecular properties, allowing it to bypass the large compute requirements of typical GNNs and function effectively with up to 100x fewer trainable parameters. 

    \item \textbf{TargetDiff}: TargetDiff \citep{guan2023model} is a 3D equivariant conditional diffusion framework that designs target-aware molecules by learning a joint generative process for both continuous atom coordinates and categorical atom types simultaneously. The model employs an SE(3)-equivariant discrete-time Markov chain to preserve essential structural symmetries. 
\end{itemize}

\subsection{Experimental Setup}

Experiments were conducted on the Bridges-2 cluster at the Pittsburgh Supercomputing Center using NVIDIA V100 (32~GB) GPUs in the
\texttt{GPU-shared} partition. For precise comparison across models, we sampled exactly one ligand per target
(\texttt{samples\_per\_pocket}~$=$~1) using each model's officially released pretrained weights. DiffSBDD used the \texttt{crossdocked\_fullatom\_cond} checkpoint at the model's default 1{,}000
diffusion steps with \texttt{min\_ligand\_nodes}~$=$~5. SimpleSBDD used its upstream scoring and center-of-mass models with an 8{,}192-SMILES ZINC pool, sampling from the 5\textsuperscript{th}--10\textsuperscript{th} percentile
score window of each pocket (seed~0). TargetDiff used the \texttt{pretrained\_diffusion} checkpoint obtained from a Zenodo mirror\footnote{\url{https://zenodo.org/records/14041881}, as the original link
was no longer available.} at \texttt{num\_steps}~$=$~1{,}000 with the upstream default seed~2021. AutoDock Vina 1.2.5 was invoked with \texttt{-{}-exhaustiveness 1 -{}-num\_modes 1}. SCASA was run with
\texttt{distance}~$=$~8.0~\AA{} and \texttt{density}~$=$~1.5.

\section{Results and Discussion}

\begin{table*}[h]
    \centering
    \caption{Performance of generative models on the control dataset (lower Vina affinity scores indicate stronger predicted binding).} 
    \label{tab:results_control}
    \begin{tabular}{lcccc}
        \toprule
        \textbf{Generative Model} & \textbf{Chemical Validity} & \textbf{Mean QED} & \textbf{Mean Sc} & \textbf{Mean Vina (kcal/mol)} \\
        \midrule
        DiffSBDD & 98.55\% & 0.2499 & 0.0022 & -5.40 \\      
        SimpleSBDD & 85.93\% & 0.6157 & 0.1674 & -7.52 \\
        TargetDiff & 87.51\% & 0.5167 & 0.6367 & -7.33 \\
        \bottomrule
    \end{tabular}
\end{table*} 

\begin{table*}[h]
    \centering
    \caption{Performance of generative models on ShallowBench (lower Vina affinity scores indicate stronger predicted binding).} 
    \label{tab:results_shallowbench}
    \begin{tabular}{lcccc}
        \toprule
        \textbf{Generative Model} & \textbf{Chemical Validity} & \textbf{Mean QED} & \textbf{Mean Sc} & \textbf{Mean Vina (kcal/mol)} \\
        \midrule
        DiffSBDD & 98.05\% & 0.2484 & -0.0030 & -4.75 \\
        SimpleSBDD & 84.97\% & 0.6167 & 0.0374 & -6.47 \\
        TargetDiff & 79.71\% & 0.4911 & 0.6088 & -5.26 \\
        \bottomrule
    \end{tabular}
\end{table*}

The control dataset results are shown in Table~\ref{tab:results_control}, and the ShallowBench results are shown in Table~\ref{tab:results_shallowbench}. We make multiple observations based on the data: 

\begin{itemize}
    \item \textbf{Systematic Decline in Binding Affinity:} Flat pockets are consistently more difficult to dock to than random control pockets across all evaluated models. For example, TargetDiff's mean Vina affinity weakens from -7.33 kcal/mol on the control dataset to -5.26 kcal/mol on ShallowBench. Similarly, SimpleSBDD experiences a drop in affinity from -7.52 kcal/mol on the control dataset to -6.47 kcal/mol on the shallow targets.
    
    \item \textbf{Trade-offs in SimpleSBDD:} SimpleSBDD achieves the strongest Vina affinities across the board (-6.47 kcal/mol on ShallowBench, -7.52 kcal/mol on the control) but at a noticeable cost to Shape Complementarity (Sc) (0.0374 and 0.1674, respectively) and chemical validity ($\sim$85\%). Because it operates more as a drug-repurposing scoring filter and COM placement baseline, pulling from drug-like ZINC libraries rather than acting as a purely de novo 3D generator, it maintains the highest QED scores ($\sim$0.62). This serves as an interesting contrast against the de novo 3D models. 
    
    \item \textbf{DiffSBDD's Weak Conditional Signal:} While DiffSBDD successfully generated the highest volume of chemically valid molecules (98.05\% on ShallowBench and 98.55\% on Control), these generations suffer from the lowest QED scores ($\sim$0.25) and a near-zero Sc. This suggests that the conditional signal directing the generation toward the pocket geometry is weak, allowing the model to produce chemically valid but largely unoptimized or untargeted structures.
    
    \item \textbf{Degradation of TargetDiff's Chemical Validity:} Unlike DiffSBDD, TargetDiff's ability to produce chemically valid molecules degrades sharply when moving from the standard control dataset (87.51\%) to the flat surfaces of ShallowBench (79.71\%). This indicates that the diffusion process TargetDiff uses struggles to enforce fundamental molecular assembly when the surrounding 3D protein constraints are sparse.
    
    \item \textbf{Shape Complementarity (Sc):} As expected due to the strict low-concavity data curation, Sc scores are generally quite low on ShallowBench. Interestingly, while DiffSBDD and SimpleSBDD drop to near-zero Sc scores on the shallow surfaces, TargetDiff still maintains a relatively high Sc of 0.6088 (compared to 0.6367 on the control). This suggests TargetDiff attempts to conform to the surface topology more aggressively than the other architectures, even if it sacrifices chemical validity to do so.
\end{itemize}   

\section{Directions for Future Work}
To advance the capability of generative models for shallow-pocket targets, there are several directions for future work.

\begin{itemize}
\item \textbf{Training Data:} Standard SBDD datasets are skewed towards high-concavity sites, causing models to learn biased distributions. To counter this, future training data should leverage more diverse datasets, potentially paired with a curriculum learning schedule that gradually introduces targets with decreasing concavity.

\item \textbf{Loss Functions:} Because flat interfaces provide minimal 3D constraints, current generative models can produce coordinates that "float" away from the protein. Addressing this may require integrating surface-vector penalties directly into the generative loss function. 

\item \textbf{Architectural Innovations:} Advancing generative modeling for shallow-pocket targets will likely require exploring new architectural modifications. To navigate non-traditional binding sites, future architectures may need to shift away from reliance on localized spatial enclosures. For example, more SBDD models may need to incorporate multi-scale graph networks capable of modeling molecular interactions across both enclosed pockets and broad planar surfaces. 

\item \textbf{Reinforcement Learning (RL):} RL could be used to optimize models for the unique physical constraints of shallow pockets. The reward function could be a weighted sum of predicted binding affinity from AutoDock Vina, shape complementarity score from SCASA, and QED score. This approach would shift the model's objective from mimicking training distributions to actively exploring new drug designs that fit best to the sparse contact areas of targets. Such RL-driven fine-tuning could explicitly teach the generator how to anchor coordinates effectively, ensuring that generated molecules do not "float" away from the interface but instead establish robust interactions. 

\item \textbf{Agentic AI:} Future work in this domain should explore the development of systems that utilize generative architectures as specialized tools within a looping workflow, iteratively refining ligands to better dock to shallow pockets. In such a framework, an AI agent could orchestrate an iterative refinement loop by sequentially calling upon external computational tools. After an initial generative pass, the agent could utilize scoring functions like AutoDock Vina for binding affinity or SCASA for shape complementarity  to evaluate the candidate's viability on the flat surface. Based on this automated feedback, the agent could iteratively prompt molecular editing models to mutate specific functional groups or extend the ligand's structural footprint to maximize interaction on sparse contact areas. This tool-augmented approach may successfully circumvent the current limitations of zero-shot generation on low-concavity interfaces.  
\end{itemize}

\section{Limitations}
While our work provides an important step toward benchmarking generative models on historically challenging low-concavity targets, we still recognize limitations. We were not able to run our evaluation on multiple random seeds or report error bars due to compute constraints. Running the heavy models and software like AutoDock Vina was already quite computationally heavy; we did not have the compute access for multiple runs. Additionally, we recognize that although AutoDock Vina is a popular computational scoring software, it may not perfectly correlate with true experimental binding affinities. 

\section{Conclusion}
In this work, we introduced ShallowBench, a rigorously curated benchmark consisting of 5,780 challenging shallow-pocket targets derived from the comprehensive CrossDocked2020 dataset. To accurately isolate these low-concavity interfaces, we developed a two-step volumetric approach that defined pocket concavity as the spatial difference between an Alpha Shape mesh "lid" and the volume occupied by the underlying protein atoms. By applying a strict upper bound on this concavity while maintaining a minimum surface area threshold, we successfully identified targets that lack traditional structural enclosures but retain enough targetable space for potential ligand interactions. Furthermore, we carefully partitioned the dataset into 4,995 training targets and 785 test targets to provide a valuable resource for researchers who would like to fine-tune models on these challenging targets. To prevent data leakage and ensure rigorous evaluation, this split was executed using 30\% sequence identity clustering, guaranteeing that no test target shares significant homology with any protein in the training set.  

By evaluating various generative models (DiffSBDD, SimpleSBDD, and TargetDiff) on this dataset, we observed a systematic decline in predicted Vina affinity relative to our control dataset. Our evaluation further revealed that without the geometric bounds of a deep pocket, frontier models struggle with distinct trade-offs. For instance, TargetDiff experienced a sharp degradation in its ability to produce chemically valid molecules, dropping to 79.71\% validity on flat surfaces, indicating that its diffusion process struggles to assemble molecules when 3D protein constraints are sparse. Conversely, DiffSBDD successfully generated chemically valid structures but suffered from near-zero shape complementarity, suggesting a weak conditional signal that fails to anchor the generation to the shallow pocket. ShallowBench therefore highlights the vulnerabilities of current architectures.

\bibliography{example_paper}
\bibliographystyle{icml2026}

\end{document}